# A Method for Robust Online Classification using Dictionary Learning: Development and Assessment for Monitoring Manual Material Handling Activities Using Wearable Sensors

Babak Barazendeh[1], Mohammadhussein Rafieisakhaei[2], Sunwook Kim[3], Zhenyu (James) Kong[3], *Member*, *IEEE*, Maury A. Nussbaum[3]

*Abstract*—Classification methods based on sparse estimation have drawn much attention recently, due to their effectiveness in processing high-dimensional data such as images. In this paper, a method to improve the performance of a sparse representation classification (SRC) approach is proposed; it is then applied to the problem of online process monitoring of human workers, specifically manual material handling (MMH) operations monitored using wearable sensors (involving 111 sensor channels). Our proposed method optimizes the design matrix (aka dictionary) in the linear model used for SRC, minimizing its ill-posedness to achieve a sparse solution. This procedure is based on the idea of dictionary learning (DL): we optimize the design matrix formed by training datasets to minimize both redundancy and coherency as well as reducing the size of these datasets. Use of such optimized training data can subsequently improve classification accuracy and help decrease the computational time needed for the SRC; it is thus more applicable for online process monitoring. Performance of the proposed methodology is demonstrated using wearable sensor data obtained from manual material handling experiments, and is found to be superior to those of benchmark methods in terms of accuracy, while also requiring computational time appropriate for MMH online monitoring.

*Note to Practitioners*—This paper develops a fast and robust classification method for online sensor data classification based on the dictionary learning principle. Due to its superior performance in terms of classification accuracy, computational speed, and robustness to non-Gaussian noise, it can be applied to a broad range of real-world applications, particularly for the scenarios that the presence of sensor data outliers causes practical difficulties for most of the existing classification algorithms.

B. Barazendeh is with the Department of Industrial and System Engineering, University of Southern California, Los Angeles, CA, 90089, USA

M. Rafieisakhaei is with the Department of Electrical and Computer Engineering, Texas A&M University, College Station, TX, 77840, USA.

S. Kim, Z. Kong and M. Nussbaum are with the Grado Department of Industrial and System Engineering, Virginia Tech, Blacksburg, VA, 24061, USA.

(corresponding author: B. Barazendeh, email: barazand@usc.edu).

## Nomenclature

| | |
|---|---|
| DL | Dictionary learning |
| DL-ROC | DL based robust online classification |
| GHNM | Greedy hybrid norm minimization |
| LASSO | Least absolute shrinkage and selection operator |
| MMH | Manual material handling |
| SRC | Sparse representation for classification |
| $\mathbb{C}_{m,n,L}$ | Solution space of a DL problem with the proper matrix dimensions |
| $\mathbf{D} \in \mathbb{R}^{m \times L}$ | Dictionary matrix |
| $\mathbf{D}_k \in \mathbb{R}^{m \times L_k}$ | Dictionary matrix for label $k$ |
| $\mathbf{d}_j \in \mathbb{R}^{m \times 1}$ | Column $j$ of dictionary $\mathbf{D}$ |
| $k > 0$ | Label index |
| $K > 0$ | Number of labels for classification |
| $\mathbf{E} \in \mathbb{R}^{m \times n}$ | Approximation error Matrix; Noise matrix |
| $\mathbf{e} \in \mathbb{R}^{m \times 1}$ | Approximation error vector; Noise vector |
| $f : \mathbb{R}^m \to \mathbb{R}$ | Error-fitting function |
| $\mathbf{G} \in \mathbb{R}^{n \times n}$ | Gram matrix |
| $g_{ij}$ | Element of $\mathbf{G}$ in row $i$ and column $j$ |
| $Q(\cdot, \cdot) : \mathbb{R}^{m \times L_k} \times \mathbb{R}^{m \times L_j} \to \mathbb{R}$ | A function related to the average coherency in a DL problem |
| $t$ | Iteration index |
| $t_{max}$ | Maximum number of iterations |
| $\mathbf{x} \in \mathbb{R}^{n \times 1}$ | Sparse representation vector |
| $\mathbf{X} \in \mathbb{R}^{L \times n}$ | Sparse representation matrix |
| $\mathbf{X}_k \in \mathbb{R}^{L_k \times n}$ | Sparse representation matrix for label $k$ |
| $\mathbf{x}_k \in \mathbb{R}^{n_k \times 1}$ | Sparse representation vector for label $k$ |
| $x_{k,j}$ | Element $j$ of the vector $\mathbf{x}_k$ |
| $\mathbf{x}_{k,j}^{row} \in \mathbb{R}^{1 \times n_k}$ | A row vector that represents the $j^{th}$ row of $\mathbf{X}_k$ |
| $\mathbf{y} \in \mathbb{R}^{m \times 1}$ | Testing data vector |
| $L > 0$ | Column dimension of the dictionary $\mathbf{D}$ |
| $L_k > 0$ | Column dimension of the dictionary $\mathbf{D}_k$ |
| $m > 0$ | Total number of measurements |
| $n > 0$ | Total number of data points |
| $n_k > 0$ | Total number of data points for label $k$ |
| $0 \leq \alpha \leq 1$ | Parameter of the hybrid norm |
| $\eta > 0$ | Balancing parameter between the reconstruction quality and incoherency terms in DL problem |
| $\Delta > 0$ | Pre-defined threshold |
| $\gamma > 0$ | Balancing parameter between the error and the sparsity enforcing term in DL problem |
| $\mu: \mathbb{R}^{m \times n} \to \mathbb{R}$ | Mutual coherency |
| $\mu_{\text{avg}}: \mathbb{R}^{m \times n} \to \mathbb{R}$ | Average mutual coherency |
| $\boldsymbol{\Psi} \in \mathbb{R}^{m \times n}$ | Training data matrix |
| $\boldsymbol{\Psi}_k \in \mathbb{R}^{m \times n_k}$ | Training data matrix for label $k$ |
| $\boldsymbol{\psi}_{k,j} \in \mathbb{R}^{m \times 1}$ | Column $j$ of $\boldsymbol{\Psi}_k$ |
| $\zeta > 0$ | Selected label based on the selection criterion |
| $\lVert \cdot \rVert_p, p \geq 0$ | $l_p$-norm of a vector or a matrix |
| $\lVert \cdot \rVert_{p,q}, p, q \geq 1$ | $l_{p,q}$-norm of a matrix |
| $\lVert \cdot \rVert_F$ | Frobenius (or Hilbert-Schmidt) norm of a matrix |
| $\lVert \cdot \rVert_{\text{hybrid},\alpha}$ | Hybrid norm of a vector or a matrix |

# I. INTRODUCTION

Sparse representation has drawn much attention in the recent years [1, 2, 3, 4, 5], due to its extensive applicability in a variety of areas. In image processing, channel coding, and machine learning, as examples, most relevant signals can be represented as a sparse linear combination of some specific bases [6]. Classification approaches based on sparse representation have achieved satisfying performance for high-dimensional data [7, 1]. For instance, sparse representation classification (SRC) has exhibited superior performance in real-world applications, such as online classification for controlling the quality of manufacturing products [8, 9, 10, 11]. As with related methods, however, the SRC approach also suffers from some limitations. For example, as mentioned in our previous work [12], SRC is highly sensitive to outliers in sensor data. This issue of sensitivity was addressed in [12], by developing the greedy hybrid norm minimization (GHNM) framework, based on a proposed novel hybrid norm. Apart from noise structure, the design matrix of the linear model formed by the training datasets (which is also called the dictionary) also has a great impact on the overall performance of SRC methods.

A popular means of SRC in the literature is to use raw data for creating the dictionary [13]. However, direct usage of raw data has some weaknesses, which if present will deteriorate the performance of SRC. First, since the data is not processed, there is the potential for redundancy in the training dataset. Second, as discussed in [14], raw data of different classes can be coherent. Third, the training data might also be contaminated with outliers.

Dictionary learning (DL) methods [15, 16, 17, 18] have been proposed to address the first two issues noted above. The goal of DL is to learn the essential features from sensor data and to remove redundancy in the training set. Moreover, by imposing some constraints in the learning process, the overall coherency of data between different classes can also be minimized. The DL process leads to a training dataset with reduced size, and subsequently decreases computational time. With a proper offline optimization process, the performance of online classification thus can be improved with the minimization of redundancy and coherency present in the data.

Existing DL approaches are designed for situations in which the training dataset is contaminated by Gaussian noise. Particularly, DL methods that are based on $l_2$-norm minimization can adequately handle Gaussian noise [19, 20, 21, 22]. In some real-world applications, though, the Gaussian noise assumption may not be valid. For instance, as investigated in [12], wherein sensor data were obtained from individuals who simulated handling materials manually in the workplace, the data may be corrupted with outliers due to a complex environment, for which the noise could be non-Gaussian. Consequently, existing DL methods based on a Gaussian noise assumption are not able to achieve the optimal classification performance for such applications.

To address this latter limitation, in this paper, the formulation of the hybrid norm [12] is adopted, based on which a DL-based robust online classification (DL-ROC) framework is proposed; it is then applied for the online monitoring of manual material handling activities using wearable sensor data. This new approach enables robust classification, due to its ability to handle outliers (non-Gaussian noise) effectively.

The remainder of this paper is organized as follows. Section 2 describes the research background and related work, while Section 3 explains the overall framework of the DL-ROC. Section 4 presents the method for the robust dictionary learning, and Section 5 assesses the performance of the proposed method using data obtained from wearable sensors during simulated manual material handling (MMH) tasks. Lastly, Section 6 concludes the paper.

# II. RESEARCH BACKGROUND AND RELATED WORK

With the rapid development of sensor technology in recent years, there has been a growing need to quickly and accurately analyze sensor data, and to make decisions online (potentially even in real-time). This need exists in many different industry sectors. One illustrative application domain is healthcare: for example, wearable sensors can be integrated with online decision-making algorithms to help elderly patients who need continuous care [23]. Another domain is in manufacturing, where there is an ongoing and critical need to monitor part quality using sensor data [24, 25, 26]. For workers in several domains who are engaged in manual material handling (MMH), the risks of musculoskeletal injury are relatively high and such risks are associated with specific work methods and exposure duration [27, 28]. For such a case, applications of wearable sensors for MMH online monitoring have the potential to be an effective means to monitor the status of the workers' operational conditions (e.g., physical demands imposed, performed task characteristics), based on which online decision making can be appropriately performed [29].

## A. The sparse signal reconstruction problem

In this section, we briefly review sparse signal reconstruction methods, including the general problem, and the least absolute shrinkage and selection operator (LASSO) [30] method, which are both directly related to the new approach proposed in this paper.

**Problem 1. The sparse signal reconstruction problem:** Given $\mathbf{y} \in \mathbb{R}^{m \times 1}$ and $\boldsymbol{\Psi} \in \mathbb{R}^{m \times n}$ ($m \ll n$), solve to find the vector $\mathbf{x} \in \mathbb{R}^n$ such that:

$$\mathbf{y} = \boldsymbol{\Psi}\mathbf{x} + \mathbf{e} \tag{1}$$

Here, the matrix $\boldsymbol{\Psi}$ is either a pre-specified transform matrix or is designed so that it fits some given signal examples [31]. Solving this problem is a challenging task, since the condition of $m \ll n$ causes the problem to be ill-posed [6]. One way to overcome this challenge is assuming sparsity of the vector $\mathbf{x}$, specifically that most of its elements are zero, which leads to the sparse solution problem [6] defined as follows:

Given sensor data $\mathbf{y} \in \mathbb{R}^{m \times 1}$, a training matrix $\boldsymbol{\Psi} \in \mathbb{R}^{m \times n}$, an error-fitting term $f: \mathbb{R}^m \to \mathbb{R}$, and a pre-defined threshold $\Delta > 0$, solve to find the sparse vector $\mathbf{x} \in \mathbb{R}^n$:

$$\begin{aligned} \min_{\mathbf{x} \in \mathbb{R}^n} \quad & \|\mathbf{x}\|_0 \\ \text{s.t.} \quad & f(\mathbf{y} - \boldsymbol{\Psi}\mathbf{x}) \le \Delta. \end{aligned} \tag{2}$$

Note that $\|\mathbf{x}\|_0$ in this problem counts the number of non-zero elements in $\mathbf{x}$. Unfortunately, this problem is NP-hard [32], though in recent years diverse approaches have been proposed. In [33], the above problem is converted to a convex one by changing the $l_0$-norm to $l_1$-norm and choosing $f(\mathbf{y} - \boldsymbol{\Psi}\mathbf{x}) = \|\mathbf{y} - \boldsymbol{\Psi}\mathbf{x}\|_2$. The resulting problem, referred to as the LASSO [30], is a convex problem, described next.

**Problem 2. The LASSO approach:** Given sensor data $\mathbf{y} \in \mathbb{R}^{m \times 1}$, a training matrix $\boldsymbol{\Psi} \in \mathbb{R}^{m \times n}$, an error-fitting term $f: \mathbb{R}^m \to \mathbb{R}$, and a pre-defined threshold $\Delta > 0$, solve to find the sparse vector $\mathbf{x} \in \mathbb{R}^n$:

$$\begin{aligned} \min_{\mathbf{x} \in \mathbb{R}^n} \quad & \|\mathbf{x}\|_1 \\ \text{s.t.} \quad & \|\mathbf{y} - \boldsymbol{\Psi}\mathbf{x}\|_2 \le \Delta. \end{aligned}$$

It has been shown that under some conditions [33], such as the Gaussian noise assumption, this convex optimization has the same solution as the original $l_0$-norm problem. Similar to many other convex programs [34], solving this optimization problem for large-scale problems with high-dimensional data can be time-consuming, which restricts the LASSO from application to online monitoring. As a result, there have been a variety of heuristic approaches proposed to solve the above problem, in which sparsity is enforced by limiting the number of iterations. Orthogonal matching pursuit (OMP) [19], stagewise OMP (StOMP) [35], and compressive sampling matching pursuit (CoSaMP) [36] are representatives examples. Unfortunately, though, all of these approaches assume Gaussian noise in their model formulation. For high-dimensional data that is contaminated with non-Gaussian noises, the GHNM method [12], which utilizes a novel hybrid norm as the error-fitting term, can efficiently solve this problem with performance sufficient for online applications.

### B. Sparse representation for classification (SRC) problem

SRC is one of the most efficient methods and as such is suitable for online supervised classification. Details of the SRC framework can be found in our previous work [12], and thus is only briefly described here.

*Notation:* Consider a problem with a total number of $K$ labels. Let $n = \sum_{k=1}^{K} n_k$ denote the total number of training points, $\boldsymbol{\Psi}_k := \left[\boldsymbol{\psi}_{k,1}, \boldsymbol{\psi}_{k,2}, \cdots, \boldsymbol{\psi}_{k,n_k}\right] \in \mathbb{R}^{m \times n_k}$ is a concatenation of training points for label $k \in \{1, 2, \cdots, K\}$, and $\boldsymbol{\Psi} = [\boldsymbol{\Psi}_1, \boldsymbol{\Psi}_2, \cdots, \boldsymbol{\Psi}_K] \in \mathbb{R}^{m \times n}$ represents the overall training matrix. Additionally, define the classification membership coefficients to be:

$$\mathbf{x} := [\mathbf{x}_1^\mathsf{T}, \mathbf{x}_2^\mathsf{T}, \cdots, \mathbf{x}_K^\mathsf{T}]^\mathsf{T},$$

such that $\mathbf{x}_k := [x_{k,1}, x_{k,2}, \cdots, x_{k,n_k}]^\mathsf{T}$ represents the classification membership coefficients for the label $k$.

*Representation of the data:* In SRC, it is assumed that the data belonging to a label has a strong inter-relationship with the data points in that label. In other words, sensor data $\mathbf{y}$ belonging to label $k$ can be represented by a linear combination of the data points in the same class, namely:

$$\mathbf{y} = \sum_{j=1}^{n_k} \boldsymbol{\psi}_{k,j} x_{k,j} + \mathbf{e}, \tag{3}$$

where $\boldsymbol{\psi}_{k,j} \in \mathbb{R}^{m \times 1}$ is the $j^{th}$ training data point (vector) of the same label $k$, $m$ is the number of sensors, $\mathbf{e}$ represents approximation error, and $n_k$ is the number of training points for label $k$. However, for a new sensor data $\mathbf{y}$, since the label of the data is unknown, the inter-relationship of the given data should be evaluated for all other labels as follows:

$$\mathbf{y} = \sum_{k=1}^{K} \sum_{j=1}^{n_k} \boldsymbol{\psi}_{k,j} x_{k,j} + \mathbf{e}. \tag{4}$$

*Label selection:* As a result of Eq. (4), the supervised classification task from Eq. (4) is equivalent to Eq. (1). However, as represented by Eq. (3), the membership coefficients $\mathbf{x}$ for data point $\mathbf{y}$ belonging to label $k$ are sparse, or $\mathbf{x} = [0, \cdots, 0, x_{k,1}, x_{k,2}, \cdots, x_{k,n_k}, 0, \cdots, 0]^\mathsf{T}$. Therefore, the classification task can be tackled by adopting the sparse signal reconstruction problem described in Eq. (2). Due to the existence of noise in the data, though, coefficients other than those belonging to label $k$ might still have small non-zero values. As a result, the following label selection criterion, which uses the energy of coefficients, can be used to calculate the label for the data vector $\mathbf{y}$:

$$\zeta := \operatorname*{argmax}_{1 \le k \le K} \frac{\|\mathbf{x}_k^*\|_2^2}{\|\mathbf{x}^*\|_2^2}, \tag{5}$$

where $\mathbf{x}^*$ is the sparse estimate of the the given data and $\mathbf{x}_k^*$ is a sub-vector of $\mathbf{x}^*$ that denotes the estimated coefficients corresponding to the $k^{th}$ label. As discussed above, raw data is used to create the training matrix $\boldsymbol{\Psi}$ (also called the dictionary). Subsequent sections will discuss that direct applications of raw data can reduce the SRC performance, and thus a new approach based on dictionary learning is applied.

*Mutual coherency of a dictionary:* For a given dictionary $\boldsymbol{\Psi} \in \mathbb{R}^{m \times n}$, the mutual coherency, $\mu: \mathbb{R}^{m \times n} \to \mathbb{R}$ [37] is defined as:

$$\mu(\boldsymbol{\Psi}) := \max_{1 \le i, j \le n, \; i \ne j} \|\boldsymbol{\psi}_i\| \|\boldsymbol{\psi}_j\|.$$

Define $\mathbf{G} := \boldsymbol{\Psi}^\mathsf{T}\boldsymbol{\Psi} \in \mathbb{R}^{n \times n}$ to be the Gram matrix associated with $\boldsymbol{\Psi}$. It can be shown that the above measure is equivalent to the off-diagonal element of $\mathbf{G}$ with the largest magnitude: let $\mathbf{G} := [g_{ij}]$, then $\mu(\boldsymbol{\Psi}) = \max_{1 \le i, j \le n, \; i \ne j} |g_{ij}|$. Moreover, the average mutual coherency $\mu_{\text{avg}}: \mathbb{R}^{m \times n} \to \mathbb{R}$ for the matrix $\boldsymbol{\Psi}$ is defined as:

$$\mu_{\text{avg}}(\boldsymbol{\Psi}) := \frac{\sum_{1 \leq i,j \leq n, \ i \neq j} |g_{ij}|}{\frac{n(n-1)}{2}}.$$

Therefore, $\mu(\boldsymbol{\Psi})$ and $\mu_{\text{avg}}(\boldsymbol{\Psi})$ respectively represent the highest and average correlation among all pairs of the columns of $\boldsymbol{\Psi}$. It has been shown that a high average mutual coherency reduces the performance of sparse signal reconstruction [37], which leads to low performance in signal classification performance.

As mentioned in Section 1 regarding creating the training matrix, the SRC simply uses the raw data [18]. In this situation, different labels of the data could share some levels of correlation (or coherency). If so, the sparse reconstruction problem becomes more difficult [37]. This is because, for a given sensor data, $\mathbf{y}$, which belongs to label $k$, there can be some data points in another label, $j$, that could also sparsely represent this sensor data. This correlation can deteriorate performance of the SRC. In order to address this issue, dictionary learning methods are utilized, which are described next.

### C. Dictionary learning (DL) approaches

To construct a more effective training matrix, some approaches have been proposed to reduce the redundancy that may exists in the raw data under the same label [38, 39, 40], as well as the coherency among different labels. These methods, which increase classification performance of [14], first started with the DL algorithms [31] that are explained in Problem (3). A generalization to learn effective training matrices for classification is formulated in Problem (4).

*Dictionary Learning:* Define $\boldsymbol{\Psi} \in \mathbb{R}^{m \times n}$ to be the training matrix that is a concatenation of the data points $\boldsymbol{\psi}_j \in \mathbb{R}^{m \times 1}, j \in \{1, \cdots, n\}$, or $\boldsymbol{\Psi} := [\boldsymbol{\psi}_1, \boldsymbol{\psi}_2, \cdots, \boldsymbol{\psi}_n]$, where each column of $\boldsymbol{\Psi}$ is a sensor reading. The goal of the DL [17] is to construct a dictionary with reduced size $\mathbf{D} := [\mathbf{d}_1, \mathbf{d}_2, \cdots, \mathbf{d}_L] \in \mathbb{R}^{m \times L}$, where $m < L < n$, so as to sparsely represent these data points by a sparse representation matrix, $\mathbf{X} \in \mathbb{R}^{L \times n}$.

**Problem 3. The DL problem:** For a given training data matrix $\boldsymbol{\Psi} \in \mathbb{R}^{m \times n}$, an error-fitting term $f \colon \mathbb{R}^m \to \mathbb{R}$, a lower target dimension $L$ ($L < n$), and a pre-defined threshold $\Delta > 0$, solve to find the sparse representation matrix, $\mathbf{X} \in \mathbb{R}^{L \times n}$, and the dictionary matrix $\mathbf{D} \in \mathbb{R}^{m \times L}$:

$$\min_{\mathbf{X}, \mathbf{D} \in \mathbb{C}_{m,n,L}} \|\mathbf{X}\|_0$$
$$\text{s. t.} \quad f(\boldsymbol{\Psi} - \mathbf{DX}) \leq \Delta,$$

where $\boldsymbol{\psi}_i = \mathbf{D}\,\mathbf{x}_i + \mathbf{e}_i$, $\mathbf{e}_i$ is a noise term, and $\mathbb{C}_{m,n,L} := \{\mathbf{X} \in \mathbb{R}^{L \times n}, \mathbf{D} \in \mathbb{R}^{m \times L} | \ \mathbf{d}_j^{\mathsf{T}} \mathbf{d}_j \leq 1, j \in \{1, \cdots, L\}\}$.

As discussed in [17], to prevent the elements of $\mathbf{D}$ from being arbitrarily large (which can cause arbitrarily small elements in $\mathbf{X}$), the $l_2$-norm of the columns of dictionary $\mathbf{D}$ is limited to be $\leq 1$, represented as a constraint in the solution space $\mathbb{C}$. Similar to the sparse reconstruction problem, and due to the existence of the $l_0$-norm in the cost function, this problem is NP-hard [41]. As the result, the $l_1$-norm is used as a relaxation for the $l_0$−norm.

*The Frobenius norm:* The optimization in Problem (3) is over $\mathbf{X}$ and $\mathbf{D}$ simultaneously, which is NP-hard [41]. To address this, an alternative utilizes the $L_{p,q}$-norm of a matrix $\mathbf{Z} :=$ $[z_{i,j}] \in \mathbb{R}^{m \times n}$, defined as $\|\mathbf{Z}\|_{p,q} = (\sum_{i=1}^{i=m}(\sum_{j=1}^{j=n}|z_{i,j}|)^p)^{q/p})^{1/q}$ for $p, q \geq 1$. We can use the $L_{2,2}$-norm, also referred to as the Frobenius norm or the Hilbert-Schmidt norm and denoted by $\|\mathbf{Z}\|_{\text{F}}$, to redefine the problem as provided subsequently.

**Problem 4. The DL problem with Frobenius norm:** Given the definitions of variables and parameters as in Problem (3), redefine the DL problem as follows:

$$\min_{\mathbf{X}, \mathbf{D} \in \mathbb{C}_{m,n,L}} \|\mathbf{X}\|_{1,1}$$
$$\text{s. t.} \quad \|\boldsymbol{\Psi} - \mathbf{DX}\|_{\text{F}}^2 \leq \Delta.$$

*Convex subproblems to solve the DL problem:* Although Problem (4) is not NP-hard, it is non-convex [41]. A common approach for solving this problem is reformulating and changing it into two subproblems [13], specifically as:

$$\min_{\mathbf{X}, \mathbf{D} \in \mathbb{C}_{m,n,L}} \|\boldsymbol{\Psi} - \mathbf{DX}\|_{\text{F}}^2 + \gamma \|\mathbf{X}\|_{1,1},$$

where $\gamma > 0$ is a balancing parameter between the error and the sparsity enforcing term. Note that the above problem is convex when the optimization is over just one variable ($\mathbf{D}$ or $\mathbf{X}$). Therefore, it is solved by fixing one variable and optimizing the other until the stopping criteria is met, which is either a maximum number of iterations or is determined when the change in cost function is less than a threshold value.

*Reducing the label coherency in classification problems:* Having a lower reconstruction error is necessary but not sufficient to obtain better classification performance. This is because certain data sets under different labels (classes) can share coherency that could significantly deteriorate classification performance [39]. There have been some efforts to improve the performance of the learnt dictionary in classification which attempted to address this issue. One approach, discussed in [14], reformulates the DL problem by adding a new term to the cost function as described subsequently.

**Problem 5. The DL problem for classification:** For a problem with $K$ labels, given the training sensor data matrix $\boldsymbol{\Psi}_k \in \mathbb{R}^{m \times n_k}, k \in \{1, 2, \cdots, K\}$, an error-fitting term $f \colon \mathbb{R}^m \to \mathbb{R}$, lower dimension sizes for each new dictionary $L_k \geq 0$, and a pre-defined threshold $\Delta > 0$, solve to find the sparse representation matrix, $\mathbf{X} \in \mathbb{R}^{L \times n}$, and the dictionary matrices $\mathbf{D}_k \in \mathbb{R}^{m \times L_k}$:

$$\min_{\{\mathbf{X}_k, \mathbf{D}_k \in \mathbb{C}_{m,n_k,L_k}, \ k=1,\cdots,K\}} \sum_{k=1}^{k=K} (\|\boldsymbol{\Psi}_k - \mathbf{D}_k \mathbf{X}_k\|_{\text{F}}^2 + \gamma \|\mathbf{X}_k\|_{1,1}$$
$$+ \eta \sum_{1 \leq j \leq K, j \neq k} \|\mathbf{D}_k^{\mathsf{T}} \mathbf{D}_j\|_{\text{F}}^2),$$

where $\eta > 0$ is a balancing parameter between the reconstruction quality and incoherency for dictionaries of different classes. The first summation in the cost function above is the same as for regular dictionary learning, where each dictionary is learned for the data of its own class. The new term of the cost function is added to decrease the level of coherency between dictionaries of different labels. This is done because the second summation leads reduced the average coherency among the data of different labels, and the above optimization problem penalizes it to have a smaller value. Therefore, the dictionary learned from the above optimization problem has better classification performance.

In the DL problem, similar to the SRC problem, if the data is contaminated by Gaussian noise, which is the case in most of the mentioned literature, the $l_2$-norm is efficient for reconstruction and learning purposes. However, we consider real-world applications, such as wearable sensors for monitoring of manual material handling, where the sensor data could be contaminated with outliers. In this situation, we adopt the idea of a hybrid norm in the reconstruction term to improve the quality of the learnt dictionary as well as the classification performance for non-Gaussian noise structures, which is discussed next.

## III. THE DL ROBUST ONLINE CLASSIFICATION (DL-ROC) APPROACH

The proposed DL-ROC framework is based on the idea of the SRC [1], which uses the general inter-relationship inside the label of the classes. However, data under different labels might have some relationship that could reduce SRC performance. Therefore, there is a need to design an effective framework to address this adverse impact on performance, by decreasing the inter-relationships among the data under different labels.

### A. The SRC Framework

As mentioned in Problem (2), when raw data is used as the training set then the existence of relationships (correlations) between the data points of different labels can reduce classification accuracy. Integrating the DL-based approach with our previously proposed framework [12] can reduce the relationships among data of different labels, which leads to a better sparse signal reconstruction and higher classification performance with smaller training sets. This new framework is represented conceptually in Figure 1. However, since in real-world applications the data can be contaminated with outliers or mixture noises, there is a need to develop a robust approach for DL problems as is discussed in the next section.

### B. Generalization of the $l_1 \oplus l_2$ hybrid norm for robust DL-based classification

In this section, the idea of the hybrid norm is generalized for DL problems. In particular, the hybrid norm for a given $\mathbf{R}$ matrix is defined as:

$$\|\mathbf{R}\|_{\mathrm{hybrid},\alpha} := \alpha\|\mathbf{R}\|_{2,2}^2 + (1-\alpha)\|\mathbf{R}\|_{1,1}$$
$$= \alpha\|\mathbf{R}\|_F^2 + (1-\alpha)\|\mathbf{R}\|_{1,1}. \qquad (6)$$

As mentioned in Section 2.2, the general DL problem is a joint optimization problem defined as [31]:

$$\min_{\mathbf{X},\mathbf{D}\in\mathbb{C}_{m,n,L}} f(\mathbf{\Psi} - \mathbf{DX}) + \gamma\,\|\mathbf{X}\|_{1,1}, \qquad (7)$$

---

**Algorithm 1. The Online Classification Framework**

*Phase I: Offline Process.*

1. Formulate the dictionary learning problem as an underdetermined system of linear equations from sensor data, $\mathbf{\Psi} = \mathbf{DX} + \mathbf{E}$;

2. Form the matrix $\mathbf{\Psi}_k = [\boldsymbol{\psi}_{k,1}, \boldsymbol{\psi}_{k,2}, \cdots, \boldsymbol{\psi}_{k,n_k}] \in \mathbb{R}^{m\times n_k}$ by randomly sampling $n_k$ training data points from the set of data belonging to label $k$ ($\forall k = 1, \cdots, K$);

3. Construct the training matrix $\mathbf{\Psi}$ by concatenating the $\mathbf{\Psi}_k$s for all $K$ labels, $\mathbf{\Psi} = [\mathbf{\Psi}_1, \mathbf{\Psi}_2, \cdots, \mathbf{\Psi}_K]$;

4. Normalize the columns of matrix $\mathbf{\Psi}$ to have a unit norm;

5. Learn the robust dictionary from $\mathbf{\Psi}$. Output: $\mathbf{D} = [\mathbf{D}_1, \mathbf{D}_2, \cdots, \mathbf{D}_K]$;

*Phase II: Online Process: Classification Membership Analysis.*

1. For a given test sample $\mathbf{y}$, approximate its sparse estimation $\mathbf{x}^*$:

$$\mathbf{x}^* := \min_{\mathbf{x}\in\mathbb{R}^n} \|\mathbf{x}\|_0, \quad \text{subject to: } f(\mathbf{y} - \mathbf{Dx}) \le \Delta;$$

2. Let $\mathbf{x}_k^*$ be the estimate of coefficient membership classification associated with label $k$, and perform classification membership analysis by setting $\zeta = \underset{1\le k\le K}{\operatorname{argmax}} \frac{\|\mathbf{x}_k^*\|_2^2}{\|\mathbf{x}^*\|_2^2}$. Output: $\zeta$.

**Figure 1**: The Online Classification Framework.

---

where $\mathbf{\Psi}$ and $\mathbf{X}$ are as in Problem (3). For learning the dictionary, and based on the Gaussian noise assumption, the $l_2$-norm is chosen as the fitting term and leads to the following problem:

$$\min_{\mathbf{X},\mathbf{D}\in\mathbb{C}_{m,n,L}} \|\mathbf{\Psi} - \mathbf{DX}\|_{2,2}^2 + \gamma\|\mathbf{X}\|_1.$$

However, based on our previous work [7], the hybrid norm is adopted for the DL problem as follows:

$$\min_{\mathbf{X},\mathbf{D}\in\mathbb{C}_{m,n,L}} \alpha\|\mathbf{\Psi} - \mathbf{DX}\|_F^2 + (1-\alpha)\|\mathbf{\Psi} - \mathbf{DX}\|_{1,1} + \gamma\|\mathbf{X}\|_{1,1}.$$

### C. Generalization of the robust dictionary learning for classification

For the SRC problem with $K$ labels, a new term $Q(\cdot,\cdot)$ is added to cost function of the DL optimization problem to improve the classification accuracy and increase incoherence among data of different classes:

$$\min_{\{\mathbf{X}_k,\mathbf{D}_k\in\mathbb{C}_{m,n_k,L_k},\ k=1,\cdots,K\}} \sum_{k=1}^{k=K}(f(\mathbf{\Psi}_k - \mathbf{D}_k\mathbf{X}_k) + \gamma\|\mathbf{X}_k\|_{1,1} + \eta\sum_{1\le j\le K, j\ne k} Q(\mathbf{D}_k^{\mathrm{T}}\mathbf{D}_j)).$$

In Ref. [14], $f(\mathbf{Y}_k - \mathbf{\Psi}_k\mathbf{X}_k) = \|\mathbf{\Psi}_k - \mathbf{D}_k\mathbf{X}_k\|_F^2$, and $Q(\mathbf{D}_k, \mathbf{D}_j) = \|\mathbf{D}_k^{\mathrm{T}}\mathbf{D}_j\|_F^2$ have been used. The fitting term $f(\cdot)$ represents the error term of data fitting in different classes, while the added term $Q(\cdot,\cdot)$ tries to reduce the average coherency among the data of different classes. Without $Q(\cdot,\cdot)$, the learned dictionaries for different classes could have high average coherency that could reduce the overall reconstruction quality, and subsequently deteriorate classification performance [19, 23]. Here, the hybrid norm [7] is used in the above equation, which leads to the following joint optimization problem for the DL-based classification:

$$\min_{\{\mathbf{X}_k, \mathbf{D}_k \in \mathbb{C}_{m,n_k,L_k}, \ k=1,\cdots,K\}} \sum_{k=1}^{k=K} (\alpha \|\mathbf{\Psi}_k - \mathbf{D}_k\mathbf{X}_k\|_F^2$$
$$+ (1-\alpha)\|\mathbf{\Psi}_k - \mathbf{D}_k\mathbf{X}_k\|_{1,1} + \gamma \|\mathbf{X}_k\|_{1,1}$$
$$+ \eta \sum_{1 \le j \le K, j \ne k} \|\mathbf{D}_k^{\mathrm{T}}\mathbf{D}_j\|_F^2). \qquad (8)$$

As mentioned in Section 2, in dictionary learning problems, where the error fitting term can take $l_1$, $l_2$ or hybrid norms, the problem is non-convex since there is a joint minimization over $\mathbf{D}$ and $\mathbf{X}$ [14]. A common method to solve this problem is to reformulate it into two steps, where each step solves the minimization over one of the variables. Therefore, the following two steps are proposed. First, initialize $\mathbf{D}^0 = [\mathbf{D}_1^0, \mathbf{D}_2^0, \cdots, \mathbf{D}_K^0]$. Then, iterate over Steps 1 and 2 described below for $t > 1$, until reaching a maximum iteration number $t_{max}$. For initiating the dictionary for each label, we could sample random from the raw data (with the same size needed for each label), or we could initiate each of them with the K-SVD method [31]. We have used the raw data for initialization.

### 1) Step 1 of the proposed DL:

In this first step, the dictionary is fixed, and the sparse representation matrix is updated:

$$\forall k \in \{1, \cdots, K\}, \mathbf{X}_k^t := \underset{\mathbf{X}_k}{\arg\min} \ (\alpha \|\mathbf{\Psi}_k - \mathbf{D}_k^{t-1}\mathbf{X}_k\|_F^2 +$$
$$(1-\alpha)\|\mathbf{\Psi}_k - \mathbf{D}_k^{t-1}\mathbf{X}_k\|_{1,1} + \gamma \|\mathbf{X}_k\|_{1,1}),$$

where $\mathbf{X}_k^t$ is the matrix $\mathbf{X}_k$ at iteration $t$, and $\mathbf{D}_k^{t-1}$ is $\mathbf{D}_k$ at iteration $t-1$ of the algorithm. In the above minimization, each column of the matrix $\mathbf{X}_k$ ($\forall k \in \{1, \dots, K\}$) can be updated independently since the dictionary is fixed. This can also be verified by the fact that the optimization cost can be separated in terms of the columns of the matrix $\mathbf{X}_k$ and their corresponding data in matrix $\mathbf{\Psi}_k$. Therefore, Step 1 can be further simplified for all $k \in \{1, \cdots, K\}$ as:

$$\forall j \in \{1, \cdots, n_k\}, \ \mathbf{x}_{k,j}^t := \underset{\mathbf{x}_{k,j}}{\arg\min} \ (\alpha \|\mathbf{\psi}_{k,j} - \mathbf{D}_k^{t-1}\mathbf{x}_{k,j}\|_F^2 +$$
$$(1-\alpha)\|\mathbf{\psi}_{k,j} - \mathbf{D}_k^{t-1}\mathbf{x}_{k,j}\|_{1,1} + \gamma \|\mathbf{x}_{k,j}\|_{1,1}),$$

where $\mathbf{x}_{k,j}$ represents the $j^{th}$ column of the matrix $\mathbf{X}_k$. This is the sparse signal reconstruction problem, and it can be solved using an approach described in our previous work [12]. In particular, the proposed method only takes a stopping criterion, which is the number of iterations or the number of residual constraints.

### 2) Step 2 of the proposed DL:

In this second step, the sparse matrix $\mathbf{X}$ is fixed and the dictionary $\mathbf{D}$ is updated:

$$\{\mathbf{D}_1^t, \cdots, \mathbf{D}_K^t\} := \underset{\{\mathbf{D}_k \in \mathbb{C}_{m,n_k,L_k}, \ k=1,\cdots,K\}}{\arg\min} \sum_{k=1}^{k=K} (\alpha \|\mathbf{\Psi}_k - \mathbf{D}_k\mathbf{X}_k^t\|_F^2$$
$$+ (1-\alpha)\|\mathbf{\Psi}_k - \mathbf{D}_k\mathbf{X}_k^t\|_{1,1} + \gamma \|\mathbf{X}_k^t\|_{1,1}$$
$$+ \eta \sum_{1 \le j \le K, j \ne k} \|\mathbf{D}_k^{\mathrm{T}}\mathbf{D}_j\|_F^2).$$

A combinatorial approach is used to solve the optimization problem in Step 2. This approach is a combination of the block coordinate descent method [42, 43] and a derivative-free optimization tool, which is the random search approach [44, 45].

*Block coordinate descent method*: In this method, each column is updated independently. Let $L_i$ be the number of columns of the $k^{th}$ dictionary $\mathbf{D}_k$. Then, in order to update the $l^{th}$ column of $\mathbf{D}_k$, or $\mathbf{d}_{k,l}$ for $l \in \{1, \cdots, L_k\}$, we can rewrite $\mathbf{\Psi}_k - \mathbf{D}_k\mathbf{X}_k$ as:

$$\mathbf{\Psi}_k - \mathbf{D}_k\mathbf{X}_k = \left(\mathbf{\Psi}_k - \sum_{j=1, j\ne l}^{L_k} \mathbf{d}_{k,j}\,\mathbf{x}_{k,j}^{\mathrm{row}}\right) - \mathbf{d}_{k,l}\mathbf{x}_{k,l}^{\mathrm{row}},$$

where $\mathbf{x}_{k,j}^{\mathrm{row}} \in \mathbb{R}^{1 \times n_k}$ is a row vector that represents the $j^{th}$ row of $\mathbf{X}_k$, i.e., $\mathbf{X}_k = [(\mathbf{x}_{k,1}^{\mathrm{row}})^{\mathrm{T}}, \cdots, (\mathbf{x}_{k,L_k}^{\mathrm{row}})^{\mathrm{T}}]^{\mathrm{T}}$.

Now, for optimizing $\mathbf{d}_{k,l}$ in Step 2, define $\mathbf{\Psi}_k^{t,l} := \mathbf{\Psi}_k - \sum_{j=1}^{l-1} \mathbf{d}_{k,j}^t \mathbf{x}_{k,j}^{t,\mathrm{row}} - \sum_{j=l+1}^{L_k} \mathbf{d}_{k,j}^{t-1} \mathbf{x}_{k,j}^{t,\mathrm{row}}$, where $\mathbf{x}_{k,j}^{t,\mathrm{row}} \in \mathbb{R}^{1 \times n_k}$ is the $k^{th}$ row of $\mathbf{X}_k^t$. Hence, for Step 2 we can solve the following optimization problem starting from $l = 1$ and increasing iterating until $l = L_k$:

$$\mathbf{d}_{k,l}^t := \underset{\mathbf{d}}{\arg\min} \left( \alpha \|\mathbf{\Psi}_k^{t,l} - \mathbf{d}\mathbf{x}_{k,l}^{t,\mathrm{row}}\|_F^2 \right.$$
$$+ (1-\alpha)\|\mathbf{\Psi}_k^{t,l} - \mathbf{d}\mathbf{x}_{k,l}^{t,\mathrm{row}}\|_{1,1}$$
$$+ \eta \sum_{1 \le j \le k-1} \|\mathbf{d}^{\mathrm{T}}\mathbf{D}_j^t\|_2^2$$
$$+ \left. \eta \sum_{k+1 \le j \le K} \|\mathbf{d}^{\mathrm{T}}\mathbf{D}_j^{t-1}\|_2^2 \right).$$

This optimization problem is convex, which can thus be solved with any derivative-free method; we use the random search approach [46, 47, 48]. Under a small enough step size, the method is shown to converge to a stationary point [49].

The proposed algorithm for robust learning dictionary is summarized as Algorithm 2 as shown in Figure 2.

---

**Algorithm 2. DL based Robust Online Classification**

**Input:** Training Signal $\Psi = [\Psi_1, \Psi_2, \cdots, \Psi_K] \in \mathbb{R}^{m \times n}$, such that $\Psi_k \in \mathbb{R}^{m \times n_k}$, $n = \sum_{k=1}^{K} n_k$, Lower Dimension for each Label $k$, $L_k$, Balancing Parameters $\alpha$ and $\gamma$, Maximum Number of Iterations $t_{max}$.

**Output:** $\{\mathbf{X}_k, \mathbf{D}_k \in \mathbb{C}_{m,n_k,L_k}, \; k = 1, \cdots, K\} =$
$$\underset{\{\mathbf{x}_k, \mathbf{D}_k \in \mathbb{C}_{m,n_k,L_k}, \; k=1,\cdots,K\}}{\arg\min} (\sum_{k=1}^{k=K}(\alpha \|\Psi_k - \mathbf{D}_k \mathbf{X}_k\|_F^2 + (1-\alpha)\|\Psi_k - \mathbf{D}_k \mathbf{X}_k\|_{1,1} + \gamma \|\mathbf{X}_k\|_{1,1} + \eta \sum_{1 \le j \le K, j \ne k} \|\mathbf{D}_k^T \mathbf{D}_j\|_F^2)).$$

1.  Initiate $\mathbf{D}^0 = [\mathbf{D}_1^0, \mathbf{D}_2^0, \cdots, \mathbf{D}_K^0]$ for each $\mathbf{D}_K$;
2.  **for** $t = 1: t_{max}$
    *Step I. Updating the sparse matrix* $\mathbf{X} = [\mathbf{X}_1, \mathbf{X}_2, \cdots, \mathbf{X}_K]$
3.      **for** k= 1: K
4.          **for** $j = 1: n_k$
    $$\mathbf{x}_{k,j}^t = \underset{\mathbf{x}_{k,j}}{\arg\min} \, (\alpha \|\psi_{k,j} - \mathbf{D}_k^{t-1} \mathbf{x}_{k,j}\|_F^2 + (1-\alpha)\|\psi_{k,j} - \mathbf{D}_k^{t-1} \mathbf{x}_{kj}\|_{1,1} + \gamma \|\mathbf{x}_{k,j}\|_{1,1});$$
5.          **end**
6.      **end**
    *Step II. Updating the Dictionary* $\mathbf{D} = [\mathbf{D}_1, \mathbf{D}_2, \cdots, \mathbf{D}_K]$
7.      **for** $k = 1: K$
8.          **for** $l = 1: L_k$
    $$\Psi_k^{t,l} = \Psi_k - \sum_{j=1}^{l-1} \mathbf{d}_{k,j}^t \mathbf{x}_{k,j}^{t,row} - \sum_{j=l+1}^{L_k} \mathbf{d}_{k,j}^{t-1} \mathbf{x}_{k,j}^{t,row};$$
9.          Solve using the Random Search Algorithm:
    $$\mathbf{d}_{k,l}^t = \underset{\mathbf{d}}{\arg\min} \, (\alpha \|\Psi_k^{t,l} - \mathbf{d}\mathbf{x}_{k,l}^{t,row}\|_F^2 + (1-\alpha)\|\Psi_k^{t,l} - \mathbf{d}\mathbf{x}_{k,l}^{t,row}\|_{1,1} + \eta \sum_{1 \le j \le k-1} \|\mathbf{d}^T \mathbf{D}_j^t\|_2^2 + \eta \sum_{k+1 \le j \le K} \|\mathbf{d}^T \mathbf{D}_j^{t-1}\|_2^2);$$
10.         **end**
11.     **end**
12. **end**

---

**Figure 2:** DL based robust online classification

## IV. CASE STUDIES

In this section, classification performance of the proposed DL-ROC framework is evaluated using the same wearable sensor data for MMH online monitoring reported in Ref. [50]. Performance was quantified from the commonly used F-score for classification along with the online computational time for efficiency, and was evaluated by comparison with the benchmark methods.

### A. Set-up for the MMH experiment

In this sub-section, the experimental set-up of the MMH monitoring and the procedures for obtaining data are briefly reviewed; more details can be found in the authors' previous work [50].

To efficiently monitor and assess workers' behaviors during MMH tasks, there is a growing interest in human activity and posture monitoring, such as using wearable sensors. Using wearable sensors, along with automatic activity classification, has great potential to enable rapid and comprehensive assessment of physical demands in diverse work settings, providing detailed information on body kinematics and work demands/strategies [50]. Sensor data obtained from MMH tasks are also considered useful and representative examples, since they involve complex, non-independent, and comprehensive data structures.

During the lab-based experiment, a wearable sensor system consisting of an inertial motion capture system (MVN BIOMECH, Xsens Technologies B. V.) [26] with 17 IMUs was used to capture 3D motion data of individuals. This system monitors the kinematics of the whole-body with a sampling rate of 60Hz. There was a total of 111 sensors (37 anatomical body landmark with 3 channels each). Throughout the experiment, time-series data consisting of the anatomical landmark locations were used, which produced 100,000 data points under each MMH task [50].

Participants in the experiment performed a simulated job that consisted of eight different major MMH tasks, and the MMH tasks correspond to eight labels for classification analysis: (1) Carrying, (2) Walking, (3) Lowering to Knee height (LoK), (4) Lowering to Ground (LoG), (5) Lifting from Knee height (LK), (6) Lifting from Ground (LG), (7) Pulling, and (8) Pushing. The wearable sensor data under each label was assigned manually from video recordings of each cycle. In the experiment, participants completed four cycles of the simulated job. Each job cycle was designed to include major MMH tasks such as lifting/lowering, pushing/pulling, and carrying. The experiment was completed by 10 young and gender-balanced volunteers whose ages were 19-29. During each MMH task, participants performed the task at their preferred work speed, but were required to complete the task in 15 seconds. To minimize fatigue, they were given short resting times after 4 cycles of the MMH tasks [50].

### B. Benchmark selection

To demonstrate the performance of the proposed DL-ROC method, a comprehensive set of benchmark methods was selected for comparison of classification performance, which have been widely applied for classification analysis in the literature. Specific methods included: support vector machine (SVM) [29], neural network (NN) [28], naïve Bayes (NB) [31], quadratic discriminant analysis (QDA) [27], and k-nearest neighborhood (k-NN) [30]. Under the SRC framework, the following two approaches were also selected as benchmark methods: 1) the OMP [19]based SRC, namely, SRC(OMP), which assumes Gaussian noise of the sensor data; and 2) the SRC based on hybrid norm of noise [12], namely, SRC(GHNM), which can handle non-Gaussian noise.

Table I. Comparison of the performance for the eight predefined states in MMH tasks obtained using the proposed approach with benchmark classification approaches (computational time (CT) per sample is in seconds). The three numbers for each classification performance correspond to F-score, Recall and Precision, respectively.

| Algorithm | Carrying | Walking | LoK | LoG | LK | LG | Pulling | Pushing | Mean Values | CT/ Sample (seconds) |
|---|---|---|---|---|---|---|---|---|---|---|
| **DL-ROC (Proposed)** | 0.9772 | 0.9546 | 0.9449 | 0.9509 | 0.9770 | 0.9363 | 0.9674 | 0.9802 | **0.9611(0.0123)** | |
| | 0.9415 | 0.9611 | 0.9115 | 0.9217 | 0.9412 | 0.9318 | 0.9578 | 0.9911 | **0.9447(0.0153)** | 0.0037 |
| | 0.9587 | 0.9535 | 0.9381 | 0.9259 | 0.9685 | 0.9358 | 0.9636 | 0.9864 | **0.9538(0.0108)** | |
| SRC (GHNM) | 0.8517 | 0.8992 | 0.8799 | 0.9015 | 0.9001 | 0.8714 | 0.8517 | 0.9010 | 0.8821(0.0203) | |
| | 0.8523 | 0.9318 | 0.8609 | 0.8711 | 0.8992 | 0.9218 | 0.8715 | 0.9115 | 0.8900(0.0149) | 0.0056 |
| | 0.8532 | 0.9069 | 0.8692 | 0.8857 | 0.9001 | 0.8899 | 0.8593 | 0.9052 | 0.8837(0.0256) | |
| SRC (OMP) | 0.8415 | 0.8912 | 0.8811 | 0.8215 | 0.8901 | 0.8511 | 0.8311 | 0.9211 | 0.8661(0.0321) | |
| | 0.8217 | 0.9102 | 0.8614 | 0.8615 | 0.9011 | 0.9412 | 0.8614 | 0.9011 | 0.8824(0.0347) | 0.0019 |
| | 0.8416 | 0.9017 | 0.8765 | 0.8508 | 0.8849 | 0.8842 | 0.8520 | 0.9127 | 0.8756(0.0401) | |
| LDA | 0.7811 | 0.6511 | 0.3517 | 0.8123 | 0.6511 | 0.6911 | 0.7214 | 0.4918 | 0.6440(0.0501) | |
| | 0.8125 | 0.7512 | 0.4911 | 0.7815 | 0.6715 | 0.7020 | 0.6518 | 0.5417 | 0.6754(0.0487) | 0.0026 |
| | 0.8042 | 0.6923 | 0.4054 | 0.7941 | 0.6598 | 0.9674 | 0.6896 | 0.5207 | 0.6917(0.0524) | |
| NN | 0.9115 | 0.7314 | 0.6112 | 0.8514 | 0.4418 | 0.7520 | 0.8911 | 0.5911 | 0.7227(0.0302) | |
| | 0.8917 | 0.7020 | 0.6478 | 0.8612 | 0.5511 | 0.7014 | 0.9015 | 0.6217 | 0.7348(0.0341) | 0.0180 |
| | 0.9019 | 0.7188 | 0.6301 | 0.8611 | 0.4823 | 0.7353 | 0.8920 | 0.6042 | 0.7282(0.0378) | |
| SVM | 0.8452 | 0.6832 | 0.8446 | 0.8854 | 0.8820 | 0.8430 | 0.8775 | 0.8424 | 0.8380(0.0243) | |
| | 0.8596 | 0.5595 | 0.8488 | 0.9892 | 0.9599 | 0.8542 | 0.9466 | 0.8525 | 0.8588(0.0291) | 0.4932 |
| | 0.8296 | 0.8692 | 0.8501 | 0.8117 | 0.8132 | 0.8299 | 0.8084 | 0.8308 | 0.8304(0.0221) | |
| *k*-NN | 0.9115 | 0.8451 | 0.6518 | 0.8614 | 0.8851 | 0.7991 | 0.9120 | 0.9215 | 0.8484(0.0185) | |
| | 0.8920 | 0.8564 | 0.7125 | 0.8751 | 0.8952 | 0.8081 | 0.9054 | 0.9152 | 0.8575(0.0178) | 0.0014 |
| | 0.9052 | 0.8492 | 0.6901 | 0.8593 | 0.9002 | 0.7925 | 0.8952 | 0.9220 | 0.8517(0.021) | |
| NB | 0.3251 | 0.2564 | 0.4521 | 0.3257 | 0.2547 | 0.4001 | 0.3258 | 0.3874 | 0.3409(0.0501) | |
| | 0.2851 | 0.3150 | 0.5217 | 0.3651 | 0.1782 | 0.3965 | 0.3814 | 0.4215 | 0.3581(0.0489) | 0.0420 |
| | 0.3102 | 0.2915 | 0.4853 | 0.4501 | 0.2185 | 0.3990 | 0.3617 | 0.4120 | 0.3660(0.0491) | |

#### C. Online classification for MMH tasks and comparison with the benchmark methods

The training and testing datasets for this study were generated first from the wearable sensor data of 10 different participants in the experiment described in Sec. 4.1. In this case study, 10,000 data points were sampled for each participant from each MMH task, which have distinct labels for classification analysis. Therefore, the data set under each label consists of 100,000 (10,000×10) sensor data points for the 10 participants. Consequently, the total number of data points of the overall data set for this case study with 8 MMH tasks/labels is 800,000 (100,000×8). The case study of classification analysis follows the following three steps.

##### Step1: Generation of the data set of this study

From the above overall data set, out of the ten participants, we randomly sampled seven of them and used their data for model training. As a result, the training dataset under each label consists of 70,000 (10,000×7) data points. The data of the remaining three participants is used for testing. So, the testing data set under each label consists of 30,000 (10,000×3) sensor data point.

##### Step 2: Data preparation for one run of classification analysis

To generate the training dataset for one run of classification analysis, under each label 2,500 data points were sampled from the training dataset generated in Step 1. Since there are a total of 8 different tasks (labels), the total number of training data points under each label is $2,500 \times 8 = 20,000$. For testing data, 1,000 data points under each label were selected randomly from the testing dataset in Step 1. In this study, the training and testing data sets were created separately from different participants to test the predictive capability of the developed model for new workers' tasks. With the proposed DL-ROC method, the original dimension of each sub-dictionary is $\mathbf{D}_k \in \mathbb{R}^{111 \times 2,500}$ (with the entire dictionary is with dimension of $\mathbf{D} \in \mathbb{R}^{111 \times 20,000}$), $k \in \{1, 2, \cdots, 8\}$, where 111 is the number of sensor channels 2,500 is the number of training data points objective of dictionary learning for $\mathbf{D}_k$ was set to $L_k = 1,600$, or $\mathbf{D}_k \in \mathbb{R}^{111 \times 1,600}$. The size of sub-dictionary under each label ($L_k = 1,600$) is determined through a numerical study, which found that the size larger than 1,600 does not improve the performance of the methodology any more. Thus, the resulting dictionary has reduced dimension of $\mathbf{D} \in \mathbb{R}^{111 \times 12,800}$. Tuning parameters for all methods were optimized using cross-validation [51].

Parameters of the benchmark methods that are based on sparsity methods are as follows. SRC(GHNM) uses the stopping criterion chosen as the residual norm is $\leq 0.01$. The same stopping criterion (residual norm is $\leq 0.01$) is used by SRC(OMP). Other benchmark methods are implemented as following, QDA using empirical prior distribution, NN with one hidden layer and 45 hidden units, SVM using Gaussian kernel, k-NN using k = 8 neighbors, and NB using Gaussian distribution and empirical prior.

Parameters of the benchmark methods that are based on sparsity methods are as follows. SRC(GHNM) uses the stopping criterion chosen as the residual norm is $\leq 0.01$. The same stopping criterion (residual norm is $\leq 0.01$) is used by SRC(OMP). Other benchmark methods are implemented as following, QDA using empirical prior distribution, NN with one hidden layer and 45 hidden units, SVM using Gaussian kernel, k-NN using k = 8 neighbors, and NB using Gaussian distribution and empirical prior.

*Step 3: Classification analysis for multiple (100) runs*

To test the consistency and robustness of the proposed method, the classification analysis was performed with 100 replicates. In each replicate, Step 2 was executed independently, namely, involving the selection of data from 7 participants for generating the training dataset and subsequent use of data from the remaining 3 participants for the testing dataset. For each selected participant, his/her data in Step 1 was sampled to generate 35the training (2,500 data points) and test dataset (1,000 data points) for each replicate. The precision, recall and F-score of these classification analyses were calculated. Table 1 compares the results of the 8 different MMH tasks in the 100 replicates using the proposed DL-ROC method along with the benchmark methods.

interpretations:

(1) The proposed DL-ROC approach achieved the best classification performance among all methods. Although its computational speed was not the fastest, it is sufficient for online applications such MMH monitoring. Considering both classification performance and computational time, DL-ROC achieved the best performance.

(2) Compared to the SRC(GHNM) method [12], the proposed approach improved the F-score (from 0.8821 to 0.9611) and reduced computational time (from 0.0056 to 0.0037 sec.). This superiority is a consequence of the fact that the proposed DL-ROC: 1) minimizes redundancy and coherency of the data, and 2) reduces the size of the training dataset.

(3) The proposed DL-ROC method outperforms SRC(OMP) in terms of classification F-score (0.9611 vs. 0.8661). This benefit stems from the capability of DL-ROC for handling non-Gaussian noise and thus outliers, whereas SRC(OMP) is based on Gaussian noise. Regarding computational speed, SRC(OMP) was faster (0.0019 sec.) due to the heuristic nature of OMP [20] .

## V. CONCLUSIONS

Most existing methods for SRC use raw data directly as a training dataset. Although this approach reduces the burden for data pre-processing, unfortunately the raw data have redundancy and coherency, and are also highly sensitive to outliers, which can each deteriorate classification performance. To address these limitations, this paper presents a DL-ROC method that is based on the DL concept and the hybrid-norm of noise. In the proposed DL process, for a given set of raw data the extracted training set is optimally learned from the raw data, and this is done in such a way that the redundancy of data under the same label and the coherency of data between different labels are both minimized. To enable robustness to outliers, the hybrid norm of the noise structure was adopted and integrated with the DL method, allowing the method to handle non-Gaussian noise.

Performance of the proposed DL-ROC method was evaluated using the case of online monitoring of MMH tasks based on wearable sensors. A comparison with several other relevant methods demonstrated that DL-ROC achieved the best performance in terms of classification accuracy (with an average F-score of 0.9611), as well good computational speed (0.0037 second/sample data). This strong performance makes the proposed method a very promising means for online classification applications, particularly for scenarios in which heavy outliers and non-Gaussian noise may exist.

## REFERENCES

[1] J. Wright, A. Y. Yang, A. Ganesh, S. S. Sastry and Y. Ma., "Robust face recognition via sparse representation," *IEEE transactions on pattern analysis and machine intelligence*, 31, no. 2 (2009): 210-227..

[2] Jianchao Yang, J. Wright, T. S. Huang and Y. Ma, "Image super-resolution via sparse representation," *IEEE transactions on image processing*, pp. 19, no. 11 (2010): 2861-2873..

[3] L. Zhang, M. Yang and X. Feng., "Sparse representation or collaborative representation: Which helps face recognition?," in *Computer vision (ICCV), 2011 IEEE international conference on, pp. 471-478. IEEE, 2011.*.

[4] J. Mairal, M. Elad and G. Sapiro, "Sparse representation for color image restoration," *IEEE Transactions on image processing* , 17, no. 1 (2008): 53-69..

[5] K. Bastani, B. Barazandeh and Z. J. Kong, "Fault Diagnosis in Multistation Assembly Systems Using Spatially Correlated Bayesian Learning Algorithm," *Journal of Manufacturing Science and Engineering,* 140, no. 3 (2018): 031003..

[6] E. J. Candès and M. B. Wakin, "An introduction to compressive sampling," *IEEE signal processing magazine* , pp. 25, no. 2 (2008): 21-30..

[7] K. Huang and S. Aviyente, "Sparse representation for signal classification," *In Advances in neural information processing systems,* pp. 609-616. 2007..

[8] S. J. Pavnaskar, J. K. Gershenson and A. B. Jambekar, "Classification scheme for lean manufacturing tools," *International Journal of Production Research ,* 41, no. 13 (2003): 3075-3090..

[9] I. McCarthy, "Manufacturing classification: lessons from organizational systematics and biological taxonomy," *Integrated Manufacturing Systems,* 6, no. 6 (1995): 37-48..

[10] R. Sabbagh and F. Ameri, "A Thesaurus-Guided Text Analytics Technique for Capability-Based Classification of Manufacturing Suppliers," in *In ASME 2017 International Design Engineering Technical Conferences and Computers and Information in Engineerin.*

[11] F. Ameri and R. Sabbagh, "Digital Factories for Capability Modeling and Visualization," in *In IFIP International Conference on Advances in Production Management Systems, pp. 69-78. Springer, Cham, 2016..*

[12] B. Barazandeh, K. Bastani, M. Rafieisakhaei, S. Kim, Z. Kong and M. A. Nussbaum, "Robust Sparse Representation-Based Classification Using Online Sensor Data for Monitoring Manual Material Handling Tasks," *IEEE Transactions on Automation Science and Engineering ,* 2017.

[13] Y. Suo, M. Dao, U. Srinivas, V. Monga and T. D. Tran, "Structured dictionary learning for classification," *arXiv preprint arXiv:1406.1943 (2014)..*

[14] I. Ramirez, P. Sprechmann and G. Sapiro, "Classification and clustering via dictionary learning with structured incoherence and shared features," *In Computer Vision and Pattern Recognition (CVPR),* pp. 2010 IEEE Conference on, pp. 3501-3508. .

[15] K. K.-D. F. Murray, B. D. Rao, K. Engan, T.-W. Lee and T. J. Sejnowski., "Dictionary learning algorithms for sparse representation," *Neural computation,* 15, no. 2 (2003): 349-396..

[16] M. Yang, L. Zhang, X. Feng and D. Zhang, "Fisher discrimination dictionary learning for sparse representation," Y. "." In Computer Vision (ICCV), 2011 IEEE International Conference on, pp. 543-550. IEEE, 2011..

[17] J. Mairal, J. Ponce, G. Sapiro, A. Zisserman and F. R. Bach, "Supervised dictionary learning," *In Advances in neural information processing systems,* pp. 1033-1040. 2009..

[18] R. Rubinstein, T. Peleg and M. Elad., "Analysis K-SVD: A dictionary-learning algorithm for the analysis sparse model," *IEEE Transactions on Signal Processing ,* 61, no. 3 (2013): 661-677..

[19] J. A. Tropp and A. C. Gilbert, "Signal recovery from random measurements via orthogonal matching pursuit," *IEEE Transactions on information theory,* 53, no. 12 (2007): 4655-4666.

[20] S. S. Chen, D. L. Donoho and M. A. Saunders, "Atomic decomposition by basis pursuit," *SIAM review,* 43, no. 1 (2001): 129-159.

[21] J. A. Tropp, "Greed is good: Algorithmic results for sparse approximation," *IEEE Transactions on Information theory ,* 50, no. 10 (2004): 2231-2242..

[22] D. P. Wipf and B. D. Rao, "Sparse Bayesian learning for basis selection," *IEEE Transactions on Signal processing,* 52, no. 8 (2004): 2153-2164..

[23] M. Iosa, G. Morone, A. Fusco, M. Castagnoli, F. R. Fusco, L. Pratesi and S. Paolucci., "Leap motion controlled videogame-based therapy for rehabilitation of elderly patients with subacute stroke: a feasibility pilot pilot study," *Topics in stroke rehabilitation ,* 22, no. 4 (2015): 306-316..

[24] J. J. Dahlgaard and S. M. Dahlgaard-Park, "Lean production, six sigma quality, TQM and company culture," *The TQM magazine ,* pp. 18, no. 3 (2006): 263-281..

[25] G. Taguchi, "Introduction to quality engineering: designing quality into products and processes.," 1986.

[26] W. Jang and M. J. Skibniewski., "A wireless network system for automated tracking of construction materials on project sites," *Journal of civil engineering and management,* 14, no. 1 (2008): 11-19..

[27] M. M. Ayoub, "Manual Materials Handling: Design And Injury Control Through Ergonomics," *CRC Press,,* 1989.

[28] K. H. E. Kroemer, "Personnel training for safer material handling," *Ergonomics,* 35, no. 9 (1992): 1119-1134..

[29] R. T. Clemen and T. Reilly, "Making hard decisions with DecisionTools," *Cengage Learning,* 2013.

[30] M. A. Figueiredo, R. D. Nowak and S. J. Wright, "Gradient projection for sparse reconstruction: Application to compressed sensing and other inverse problems," *IEEE Journal of selected topics in signal processing ,* 1, no. 4 (2007): 586-597.

[31] M. Aharon, M. Elad and A. Bruckstein, "$ rm k $-SVD: An algorithm for designing overcomplete dictionaries for sparse representation," *IEEE Transactions on signal processing,* 54, no. 11 (2006): 4311-4322.

[32] W. Dai and O. Milenkovic, "Subspace pursuit for compressive sensing signal reconstruction," *IEEE Transactions on Information Theory ,* pp. 55, no. 5 (2009): 2230-2249..

[33] D. L. Donoho, "For most large underdetermined systems of linear equations the minimal ℓ1-norm solution is also the sparsest solution," *Communications on pure and applied mathematics ,* 59, no. 6 (2006): 797-829..

[34] V. Cevher, S. Becker and M. Schmidt, "Convex optimization for big data: Scalable, randomized, and parallel algorithms for big data analytics," *IEEE Signal Processing Magazine ,* 31, no. 5 (2014): 32-43.

[35] D. David L., Y. Tsaig, I. Drori and J.-L. Starck., "Sparse solution of underdetermined systems of linear equations by stagewise orthogonal matching pursuit," *IEEE Transactions on Information Theory*, 58, no. 2 (2012): 1094-1121..

[36] N. Deanna and J. A. Tropp, "CoSaMP: Iterative signal recovery from incomplete and inaccurate samples," *Applied and Computational Harmonic Analysis*, 26, no. 3 (2009): 301-321..

[37] M. Elad, "Optimized projections for compressed sensing," *IEEE Transactions on Signal Processing*, 55, no. 12 (2007): 5695-5702..

[38] Z. Jiang, Z. Lin and L. S. Davis, "Label consistent K-SVD: Learning a discriminative dictionary for recognition," *IEEE Transactions on Pattern Analysis and Machine Intelligence*, 35, no. 11 (2013): 2651-2664.

[39] S. Kong and D. Wang, "A dictionary learning approach for classification: Separating the particularity and the commonality," pp. Computer Vision–ECCV 2012 (2012): 186-199.

[40] M. Yang, D. Dai, L. Shen and L. V. Gool., "Latent dictionary learning for sparse representation based classification," *In Proceedings of the IEEE Conference on Computer Vision and Pattern Recognition*, pp. 4138-4145. 2014..

[41] M. Razaviyayn, H.-W. Tseng and Z.-Q. Luo, "Dictionary learning for sparse representation: Complexity and algorithms," *In Acoustics, Speech and Signal Processing (ICASSP)*, pp. 2014 IEEE International Conference on, pp. 5247-5251. IEEE, 2014..

[42] M. Razaviyayn, M. Hong and Z.-Q. Luo, "A unified convergence analysis of block successive minimization methods for nonsmooth optimization," *SIAM Journal on Optimization*, 23, no. 2 (2013): 1126-1153..

[43] J. Mairal, F. Bach, J. Ponce and G. Sapiro, "Online dictionary learning for sparse coding," *In Proceedings of the 26th annual international conference on machine learning*, pp. pp. 689-696. ACM, 2009..

[44] L. M. Rios and N. V. Sahinidis, "Derivative-free optimization: a review of algorithms and comparison of software implementations," *Journal of Global Optimization*, 56, no. 3 (2013): 1247-1293 .

[45] L. A. Rastrigin, "The convergence of the random search method in the extremal control of a many parameter system," *Automaton & Remote Control*, 24 (1963): 1337-1342..

[46] W. L. Price, "Global optimization by controlled random search," *Journal of Optimization Theory and Applications*, 40, no. 3 (1983): 333-348.

[47] J. Bergstra and Y. Bengio, "Random search for hyper-parameter optimization," *Journal of Machine Learning Research*, 13, no. Feb (2012): 281-305.

[48] Z. B. Zabinsky, "Random search algorithms," *Wiley Encyclopedia of Operations Research and Management Science (2009)*.

[49] F. J. Solis and R. J.-B. Wets, "Minimization by random search techniques," *Mathematics of operations research*, pp. 6, no. 1 (1981): 19-30.

[50] S. Kim and M. A. Nussbaum, "An evaluation of classification algorithms for manual material handling tasks based on data obtained using wearable technologies," *Ergonomics 57, no. 7 (2014): 1040-1051*.

[51] R. Kohavi, "A study of cross-validation and bootstrap for accuracy estimation and model selection," *Ijcai*, vol. 14, no. 2, pp. 1137-1145. 1995..